\documentclass{article}

    \usepackage[preprint]{neurips_2025}

\usepackage[utf8]{inputenc} %
\usepackage[T1]{fontenc}    %
\usepackage{hyperref}       %
\usepackage{url}            %
\usepackage{booktabs}       %
\usepackage{amsfonts}       %
\usepackage{nicefrac}       %
\usepackage{microtype}      %
\usepackage{xcolor}         %

\usepackage{multirow} %
\usepackage{graphics} %
\usepackage{graphicx} %
\usepackage{subcaption} %
\usepackage{comment} %
\usepackage[frozencache,cachedir=.]{minted} %
\title{Adding simple structure at inference improves Vision-Language Compositionality}

\author{Imanol Miranda\quad Ander Salaberria \quad Eneko Agirre \quad Gorka Azkune \\
HiTZ Center -- Ixa, University of the Basque Country (UPV/EHU) \\
\texttt{\{imanol.miranda, ander.salaberria, e.agirre, gorka.azcune\}@ehu.eus} \\
}

\begin{document}

\maketitle

\begin{abstract}
  Dual encoder Vision-Language Models (VLM) such as CLIP are widely used for image-text retrieval tasks. However, those models struggle with compositionality, showing a bag-of-words-like behavior that limits their retrieval performance. Many different training approaches have been proposed to improve the vision-language compositionality capabilities of those models. In comparison, inference-time techniques have received little attention. In this paper, we propose to add simple structure at inference, where, given an image and a caption: i) we divide the image into different smaller crops, ii) we extract text segments, capturing objects, attributes and relations, iii) using a VLM, we find the image crops that better align with text segments obtaining matches, and iv) we compute the final image-text similarity aggregating the individual similarities of the matches. Based on various popular dual encoder VLMs, we evaluate our approach in controlled and natural datasets for VL compositionality. We find that our approach consistently improves the performance of evaluated VLMs without any training, which shows the potential of inference-time techniques. The results are especially good for attribute-object binding as shown in the controlled dataset. As a result of an extensive analysis: i) we show that processing image crops is actually essential for the observed gains in performance, and ii) we identify specific areas to further improve inference-time approaches.  
  
\end{abstract}

\section{Introduction}
\label{sec:intro}
Image-text retrieval is the task where given an image and a set of texts, the relevant text has to be retrieved (or vice-versa). This is a long-studied task due to its practical applications and scientific implications \cite{ijcai2022p759}. In recent years, dual encoder Vision-Language Models (VLM), such as the pioneering CLIP \cite{radford2021learning} or the modern SigLIP 2 \cite{tschannen2025siglip}, have topped almost all the benchmarks in terms of performance and efficiency. However, those dual encoder models perform poorly when tested for Vision-Language Compositionality \cite{thrush2022winoground, hsieh2024sugarcrepe, miranda2024BiVLC}, where models are evaluated to see whether they can distinguish between images and textual descriptions that share the same elements, but in different arrangements. Dual encoder VLMs have been shown to behave similarly to bag-of-words systems \cite{yuksekgonul2022and}, struggling to distinguish between \textit{"a black cat and a white dog"} and \textit{"a white cat and a black dog"}. This limitation hampers the performance of image-text retrieval and suggests that dual encoder VLMs do not learn compositional representations.  

\begin{figure}[ht]
    \centering
    \begin{subfigure}[b]{0.42\linewidth}
         \includegraphics[width=\textwidth]{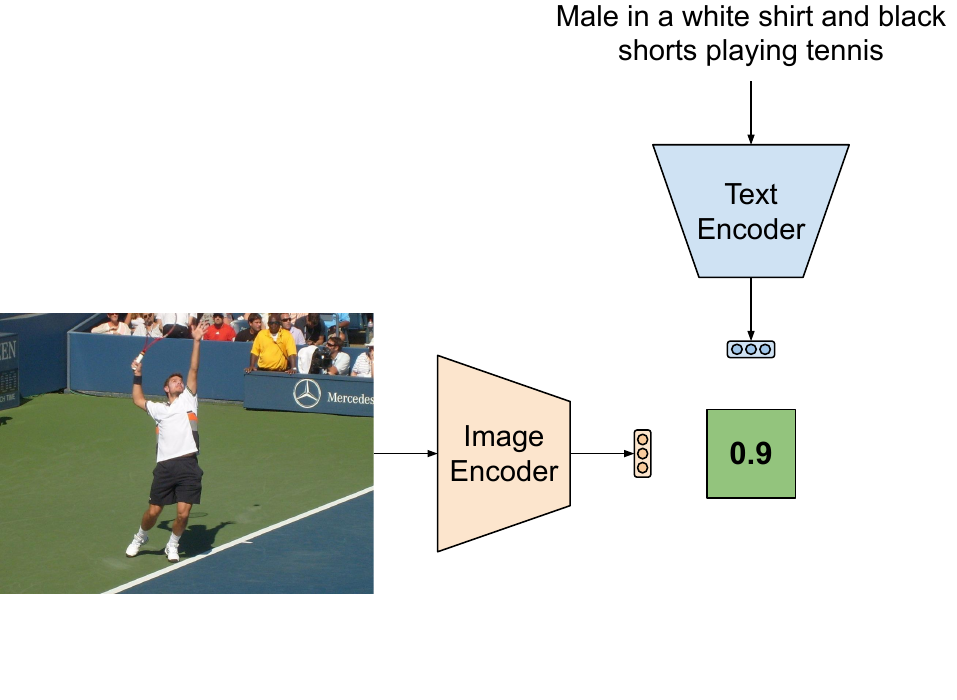}
         \caption{Baseline.}
         \label{fig:itm_baseline}
    \end{subfigure} 
    \begin{subfigure}{0.57\linewidth}
        \centering
        \includegraphics[width=\textwidth]{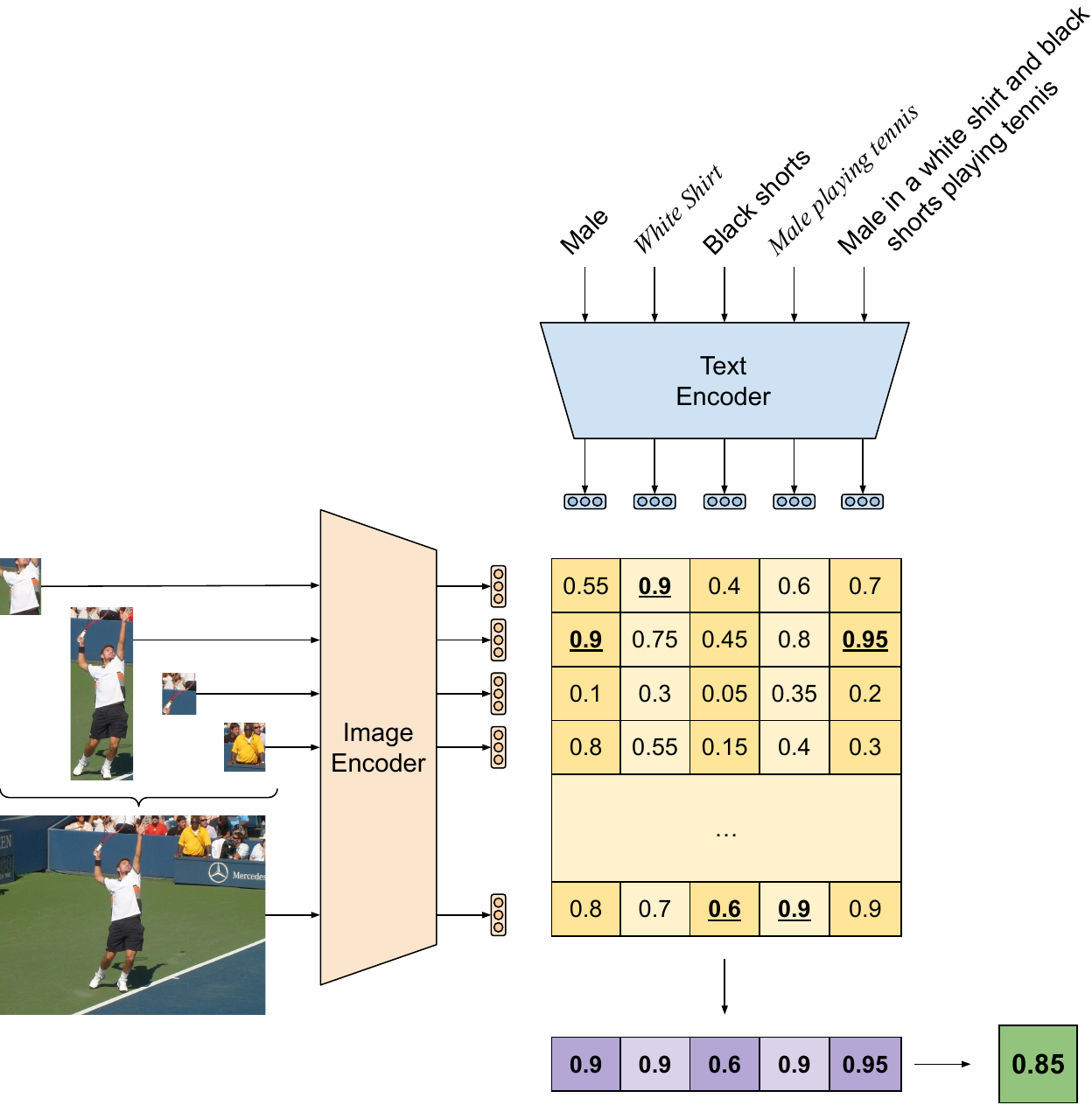}
        \caption{Our method, adding structure at inference.}
        \label{fig:itm_atoms_crops}
    \end{subfigure}
    \caption{(a) Dual-encoder VLMs compute image-text similarity as the cosine similarity between the text and image vectors. (b) We add structure at inference dividing the image in crops of different sizes (sample shown) and the text in text segments (sample shown). We compute a similarity matrix, and select the best matching crop for each text segment (underlined). The final similarity is the average of the similarity for the matches. }
    \label{fig:itm}
\end{figure}

Most of the previous work to improve VL compositionality focuses on end-to-end training approaches \cite{lewei2021filip, gao2022pyramidclip, jing2024fineclip}. Despite the good results reported on several benchmarks, end-to-end training has its own limitations. For example, \cite{lewis2024does} show the generalization problems of end-to-end training approaches when out-of-distribution compositions are evaluated, questioning whether further training actually makes the learned representations more compositional. On the other hand, \cite{campbell2024understanding} analyze the so-called \textit{binding problem} and argue that the existence of that binding problem in VLMs is an indication of using compositional representations, suggesting that pretrained VLMs may be more capable than initially evident. 

Inspired by those works, instead of focusing on end-to-end training, we explore how to add simple structure at inference to improve the VL compositionality of dual encoder VLMs. Given a pretrained VLM such as CLIP, we propose the following method to compute the similarity of an image and a text: i) we divide the image into different smaller \textbf{crops}, ii) we extract \textbf{text segments} from the captions, capturing objects, attributes and relations, iii) we find the set of best 
crop-segment \textbf{matches}, i.e. for each text segment, find the image crop with maximum similarity according to the VLM, and iv) we aggregate the similarities of every match, obtaining the final score (see Figure \ref{fig:itm}). 

We follow the evaluation procedure in the field \cite{lin2024revisiting, lin2024evaluating} and use bidirectional retrieval datasets, where each retrieval instance is formed by two images and two captions, hard negatives between them. We focus on \textsc{Swap} cases, due to two reasons: i) previous work shows that \textsc{Swap} cases are by far the hardest ones for VLMs \cite{thrush2022winoground, hsieh2024sugarcrepe, miranda2024BiVLC}, and ii) \textsc{Swap} cases use the same words for positive and negative captions, but in different arrangements, effectively testing the bag-of-words problem\footnote{Notice that other typical compositionality cases are \textsc{Replace} and \textsc{Add}. Those cases are easier for VLMs, since images and captions contain different elements.}. Current VL compositionality evaluations provide only test data, which hampers the principled development of new techniques. We thus decided to build two new development datasets, called \textsc{BiSCoR-Nat} and \textsc{BiSCoR-Ctrl}. After developing different instantiations of our inference-time approach in the proposed development datasets, we evaluate the best configurations in \textsc{BiVLC} \cite{miranda2024BiVLC}, Winoground \cite{thrush2022winoground} and the test split of \textsc{BiSCoR-Ctrl}. As a result of our experiments we find that:

\noindent \textbf{1)} Our inference-time approach significantly improves the retrieval performance of two well-known dual encoder VLMs (CLIP and SigLIP 2), showcasing the potential of inference-time approaches for VL compositionality. Those improvements are observed consistently for all evaluation datasets.

\noindent \textbf{2)} Ablation experiments show that dividing images into smaller crops or regions is actually essential for the observed performance gains. In fact, the results barely improve or even degrade when only text segments are used. 

\noindent \textbf{3)} Our inference-time approach is specially strong for attribute-object binding, i.e. a correct assignment of attributes to the relevant object, a well-known problem of VLMs \cite{greff2020binding}. In our controlled set-up, we observe improvements of up to 90 absolute points for some cases.

\noindent \textbf{4)} We provide an extensive analysis of the results, identifying specific areas to further improve our inference-time approach. We show that the calculation of the final score for image-text similarity is sensitive to certain biases and that the automatic generation of text segments can also be improved.

We publicly release the code\footnote{https://github.com/IMirandaM/structure-inference-compositionality} for all the experiments, as well as the new datasets\footnote{https://huggingface.co/datasets/imirandam/BiSCoR}. %

\section{Related work}
\label{sec:sota}

\paragraph{End-to-end training for VL compositionality.} %
In the literature, we can find two main end-to-end training paradigms to improve VL compositionality in dual encoder VLMs: i) to incorporate fine-grained alignments between text and images during training, which is implemented in several ways by X-VLM \citep{zeng2021xvlm}, PyramidCLIP \citep{gao2022pyramidclip}, FILIP \citep{lewei2021filip} and FineCLIP \citep{jing2024fineclip}; ii) using hard-negative examples to improve the contrastive learning step, as shown by NegCLIP \citep{yuksekgonul2022and}, which only uses hard-negative texts, and GNM \citep{sahin2024enhancing}, TripletCLIP \citep{patel2024TripletCLIP} or CLIP\textsubscript{TROHN-Img} \citep{miranda2024BiVLC}, which also use hard-negative images of different nature. However, end-to-end training approaches struggle to generalize to out-of-distribution compositions \citep{lewis2024does}, so we will consider other research avenues in this work.

\paragraph{Inference-time approaches.} Some works have also considered using inference-time approaches for image-text matching, usually for evaluating text-to-image generation. The most extended idea is to generate questions conditioned on the caption to then evaluate images in a VQA setting and compute similarities using VQA scores \citep{hu2023tifa}. CECE adds another step by first generating captions that entail or contradict the original one, and then using them to generate yes/no questions via templates \citep{cascantebonilla2024cece}. GECKO improves its performance by enhancing its VQA scoring and enforcing that each word of the caption is mentioned in at least one question \citep{wiles2025gecko}; and DSG takes it a step further and omits questions that do not make sense given previous responses \citep{cho2024davidsonian}.%

Our approach differs from previous works in various important aspects: i) we generate image crops and align them to text segments, instead of using only the entire image, ii) we do not use an off-the-shelf VQA system, but the dual encoder VLM itself to compute the similarity between image crops and text segments. We show that using image crops is indeed crucial for the good results we obtain.

\paragraph{Evaluation datasets.} The majority of the published evaluation datasets for VL compositionality are formulated as image-to-text retrieval tasks, where given an image and two or more captions, one positive and the others hard negatives, a system has to retrieve the positive caption. ARO \citep{yuksekgonul2022and}, CREPE \citep{ma2023crepe} and SugarCrepe \citep{hsieh2024sugarcrepe} are representative of those approaches. Despite efforts to build plausible and grammatically well-formed hard-negative captions, several works have already demonstrated that benchmarks focusing on image-to-text retrieval unintentionally reflect some language-based biases, which VLMs can take advantage of \citep{lin2024evaluating,lin2024revisiting}. Hence, bidirectional retrieval datasets are the preferred option to evaluate VL compositionality, forming each retrieval instance with two images and two captions, which are hard negatives between them. Winoground is a bidirectional retrieval dataset with 400 evaluation instances containing only \textsc{Swap} cases \citep{thrush2022winoground}. However, further analysis on the dataset has shown that compositionality is only relevant in 171 instances \citep{diwan2022winoground}. Similarly, EqBen is based on video-text datasets but suffers from low-quality images \citep{lin2024evaluating}, which were generated using a graphics engine \citep{wang2023equivariant}. A later release solved this issue, but it only contains 280 high-quality instances and does not cover \textsc{Swap} cases. BiVLC has solved the previous problems by expanding SugarCrepe with synthetic images. It contains 3K instances, which are obtained after manually filtering incorrect or ambiguous instances \citep{miranda2024BiVLC}.

\section{The Proposed Inference-Time Approach}
\label{sec:proposal}

Following some previous studies \cite{campbell2024understanding}, we assume that pretrained VLMs already learn compositional representations (to some extent). However, we hypothesize that the inference mechanism typically used - cosine similarity between a vector representing the full image and a vector representing the full text - does not leverage the underlying representations well. To validate our hypothesis, we propose a novel inference-time approach to calculate the similarity score between an image and a text, using pretrained VLMs without any further training. Our approach is based on four steps described next (see also Figure \ref{fig:itm} for a visual explanation):

\paragraph{[Step 1] Generate image crops:} Given an input image $I$, we define different crops $\{c_i\}$ where $i \in [0, N]$. The final set of crops depends on the values of the scale, aspect ratio, and stride hyperparameters. The objective of this step is to generate different image regions that may capture relevant visual aspects of the whole image.

\paragraph{[Step 2] Generate text segments:} Given a textual description $T$, we segment the text to obtain hierarchical object-attribute-relation segments $t_j$ where $j \in [0, M]$. For example, for $T=$ "a black cat and a white dog", the generated text segments could be: $t_0=$ "cat", $t_1=$ "dog", $t_2=$ "black cat", $t_3=$ "white dog" and $t_4=$ "a black cat and a white dog". The generation of text segments can vary depending on the granularity of the segments we want to have. For example, we may prefer to ignore isolated objects and use always objects with attributes. In the previous example, that would imply removing $t_0=$ "cat" and $t_1=$ "dog".

\paragraph{[Step 3] Search for matches between crops and text segments:} Assuming a pretrained dual encoder VLM with visual encoder $V_{enc}$ and textual encoder $T_{enc}$, we denote $v_i = V_{enc}(c_i)$ the vector representing image crop $c_i$ and $l_j = T_{enc}(t_j)$ the vector representing text segment $t_j$. We compute the similarity values of every crop $c_i$ with every text segment $t_j$ as $sim(c_i, t_j) = cos\_sim(v_i, l_j)$ and build a matrix of similarities of size $N \times M$. For every text segment $t_j$, we select the matching crop $c_k$ with maximum similarity to $t_j$, i.e. the \textbf{matching} pair is $m_j = (c_k, t_j)$ where $k = argmax_i \{sim(t_j, c_i)\}$. As a result of this %
process, we get a set of matches $m_j$ where $j \in [0, M]$, having one match per text segment.

\paragraph{[Step 4] Compute image-text alignment score:} The final score between image $I$ and text $T$ is the average of the similarities of the matches, i.e. $Sim(I, T) = \frac{1}{M} \sum_j sim(m_j)$.

\section{Experiments and Results}
\label{sec:experiments}
In this section we first introduce the models we use for evaluating our inference-time approach (Section \ref{subsec:models}). Our approach can be instantiated in various ways, changing the values of the cropping hyperparameters (scale, aspect ratio and stride), as well as the text segmenting strategies. To understand how those design choices affect the VL compositionality performance, we run several development experiments next (Section \ref{subsec:dev}). Using the best configuration from those experiments, we evaluate our approach in existing VL compositionality datasets (Section \ref{subsec:eval}).

\subsection{Models}
\label{subsec:models}
All our experiments are performed with dual encoder models, as they are the \textit{de facto} standard for image-text retrieval. Dual encoder VLMs learn separate encoders for images and text, aligning their embeddings in a shared space. %
In this work, we evaluate CLIP \citep{radford2021learning} and SigLIP 2 \citep{tschannen2025siglip}. We select CLIP because as a pioneering VLM, it has been widely studied. SigLIP 2, on the other hand, is a modern dual encoder VLM, which stands for its multilingual nature and combined training approach, mixing self-supervised and language-supervised training techniques. For CLIP we use the ViT-B/32 visual encoder with an input image size of $224^2$. For SigLIP 2 we evaluate the ViT-B/32 and the giant-opt-ViT/16 architectures, both with input size $256^2$.

\subsection{Development experiments}
\label{subsec:dev}

\begin{figure}[t]
  \centering \includegraphics[width=1\textwidth]{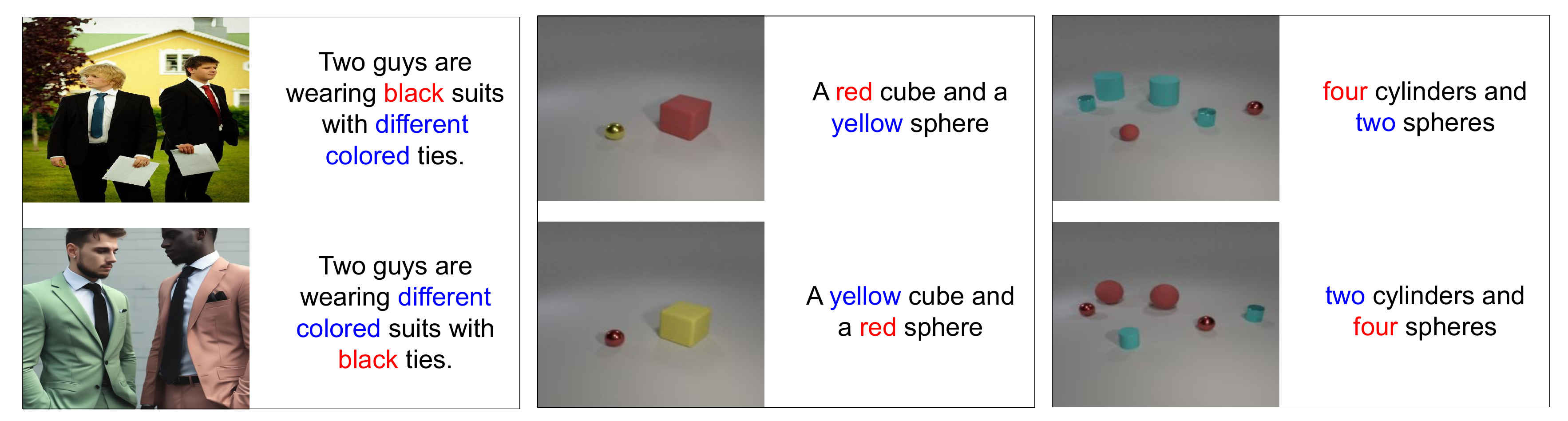}
  \caption{Instances from the \textsc{BiSCoR} development dataset. From left-to-right an instance from \textsc{BiSCoR-Nat} followed by two instances of \textsc{BiSCoR-Ctrl}, the first one for \textsc{Color} and the second for \textsc{Quantity} with 6 objects. Each instance consists of a positive pair (top image and caption) and a hard negative pair (bottom image and caption).}
  \label{fig:BiSCoR-examples}
\end{figure}

Given the lack of development datasets for VL compositionality, we build our own ones, grouped under the name of \textsc{BiSCoR}, a dataset family of Bidirectional Swaps for Compositional Reasoning development. The most important design decisions are: i) the datasets only consider the \textsc{Swap} cases, since they are the hardest cases for VLMs \cite{hsieh2024sugarcrepe, miranda2024BiVLC}; ii) to avoid any language bias, the datasets are based on bidirectional retrieval, following Winoground and \textsc{BiVLC}, i.e. each instance has two images and two captions which are hard negatives between them; iii) we use the performance metrics proposed by Winoground, i.e. image-to-text retrieval accuracy (I2T), text-to-image retrieval accuracy (T2I) and group score (Group). Formal definitions can be found in Appendix~\ref{appendix:metrics}. As group score aggregates the I2T and T2I for an instance, it will be the main metric for evaluation throughout this work.

The \textsc{BiSCoR} dataset family has a a branch with natural captions and images, \textsc{BiSCoR-Nat}, and a branch  for controlled development, \textsc{BiSCoR-Ctrl} (see Figure \ref{fig:BiSCoR-examples} and Appendix \ref{appendix:more_examples} for examples). \textsc{BiSCoR-Nat} is a subset of \textsc{TROHN-Img}, a noisy dataset presented in \citep{miranda2024BiVLC} for training purposes. Each instance consists of two images and two captions. First, we detect the \textsc{Swap} cases using a simple heuristic: we automatically check whether both captions of an instance contain exactly the same words. After applying the heuristic, we end up with nearly 6,000 potential \textsc{Swap} candidates. As the dataset is noisy, we manually filter  with three annotators all the instances whose images and captions are not well aligned, resulting in 427 high-quality \textsc{Swap} cases. See Appendix~\ref{appendix:dataset_info} for more details. 

\textsc{BiSCoR-Ctrl}, on the other hand, is designed to have full control of the scenes and their compositions, allowing for a finer-grained evaluation of compositionality. For that purpose, we base the dataset on CLEVR \citep{johnson2017clevr}, a dataset for
compositional language and elementary visual reasoning built with a 3D image rendering engine. We generate different variants of \textsc{Swap} cases: i) \textsc{Color}, ii) \textsc{Size}, iii) \textsc{Material} and iv) \textsc{Quantity}. We build 1K instances per variant based on the scenes of the training set of CLEVR. Furthermore, as we programmatically generate the images and captions, we also have the scene graphs of both, so we can apply different text segmenting strategies without any noise, faithfully assessing the merits of each text segmenting strategy. Finally, we can also ensure that positive and negative images have exactly the same appearance, avoiding any kind of bias. More details of the dataset are provided in Appendix~\ref{appendix:dataset_info}.

We evaluate six different configurations of our inference-time approach. For image crops, we always use crops of sizes (32, 32), (56, 56), (112, 112), (224, 224), (56, 112) and (112, 56), combining different scales and aspect ratios. We resize all the crops to the input size of the model and we deploy those crops in two different ways: i) \textit{grid}, avoiding any overlap of crops of the same size, and ii) \textit{overlap}, using a stride of $crop\_size / 2$. This means that we process 86 crops per image with the \textit{grid} configuration, and 270 crops with \textit{overlap}. 

\begin{figure}[ht]
    \centering 
         \includegraphics[width=0.7\textwidth]{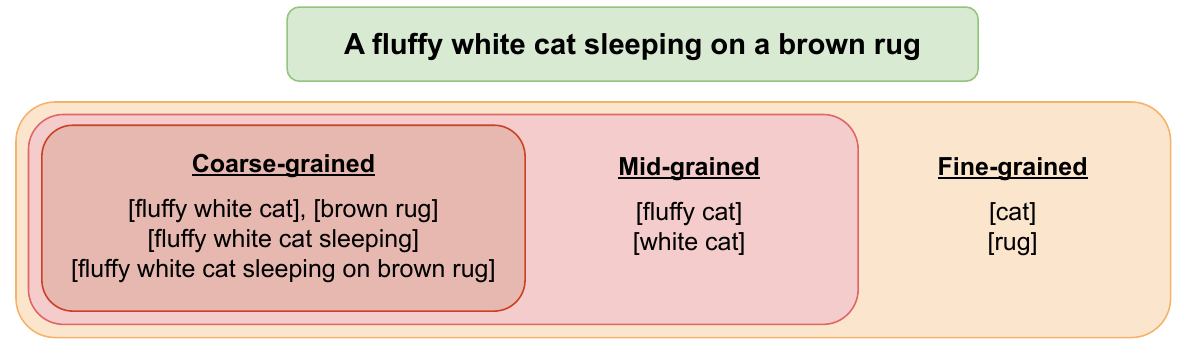}   
    \caption{An example for our three text segmenting strategies. As can be seen, \textit{mid-grained} adds two new text segments to the ones generated for \textit{coarse-grained}, and \textit{fine-grained} further adds two more segments to \textit{mid-grained} text segments.}
    \label{fig:text-segments}   
\end{figure}

Regarding text segmentation, we consider three different strategies: i) \textit{fine-grained}, where text segments are of the form [object], [attribute + object] and [attribute + object, relation, attribute + object], ii) \textit{mid-grained}, where we avoid objects in isolation, thus we have [attribute + object] and [attribute + object, relation, attribute + object], and iii) \textit{coarse-grained} text segments, which is the same as \textit{mid-grained} but for each object we include all the attributes of that object in the text segment  (see Figure \ref{fig:text-segments}). To obtain those text segments in \textsc{BiSCoR-Ctrl} we directly use the ground-truth scene graphs, so we know that the created text segments are perfect. However, for \textsc{BiSCoR-Nat}, as the captions are written by humans and we do not have any gold scene graphs, we use an LLM to extract those segments from captions (see Appendix~\ref{appendix:LLM} for more details).

We evaluate the six configurations for the three models on the development data. Table \ref{tab:dev-results} shows the results for SigLIP 2-Giant, as the other models show the same trend (see all results at Appendix~\ref{appendix:dev_results}). 

\begin{table}[t]
  \caption{Development results with SigLIP 2-Giant. FG for \textit{fine-grained}, MG for \textit{mid-grained} and CG for \textit{coarse-grained} text segments. Note that for \textsc{BiSCoR-Ctrl} the MG and CG text segments are the same. Bold for best results. }
  \label{tab:dev-results}
  \centering
  \resizebox{\columnwidth}{!}{%
  \begin{tabular}{p{2.5cm}ccccccccccccccccccc}
    \toprule
     \multirow{2}{*}{\small\textbf{Model}} &
     \multicolumn{3}{c}{\textbf{\textsc{BiSCoR-Nat}}} & \multicolumn{12}{c}{\textbf{\textsc{BiSCoR-Ctrl}}} \\
     & & & & \multicolumn{3}{c}{\textbf{\textsc{Color}}} & \multicolumn{3}{c}{\textbf{\textsc{Size}}} & \multicolumn{3}{c}{\textbf{\textsc{Material}}}& \multicolumn{3}{c}{\textbf{\textsc{Quantity}}} \\
\cmidrule(lr){2-4} \cmidrule(lr){5-7}  \cmidrule(lr){8-10}  \cmidrule(lr){11-13}  \cmidrule(lr){14-16} 
     &  \textbf{I2T} & \textbf{T2I} & \textbf{Gr.}& \textbf{I2T} & \textbf{T2I} & \textbf{Gr.}&  \textbf{I2T} & \textbf{T2I} & \textbf{Gr.} &  \textbf{I2T} & \textbf{T2I} & \textbf{Gr.}&  \textbf{I2T} & \textbf{T2I} & \textbf{Gr.} \\
    \midrule    
    Random & 25.0 & 25.0 & 16.7 & 25.0 & 25.0 & 16.7 & 25.0 & 25.0 & 16.7 & 25.0 & 25.0 & 16.7 & 25.0 & 25.0 & 16.7\\
    \midrule
    SigLIP 2-Giant & 75.4          & 9.1           & 8.2           & 19.5          & 8.7           & 4.8           & 18.7         & 15.8          & 4.4           & 15.5          & 7.8           & 2.2           & 21.7          & 14.9          & 4.9 \\
    \midrule
    \quad Grid + FG & 82.4          & 43.1          & 39.8          & 95.9          & 91.6          & 90.1         & 40.0          & 17.2          & 9.2           & 94.3          & 80.4          & 78.0          & 39.1          & 25.1          & 18.8 \\
    \quad Grid + MG &  82.0          & 50.1          & 47.3          & 95.9          & 94.0          & 92.3          & 40.0          & 26.0          & 14.9          & 94.3          & 86.7          & 84.1          & 33.6          & 24.9          & 14.9 \\
    \quad Grid + CG &  83.4 & 52.2 & 49.0 & -          & -          & - & -          & -          & -          & -          & -          & -          & -          & -          & - \\
    \midrule
    \quad Over + FG & \textbf{84.3} & 45.9          & 43.6          & \textbf{97.6} & 94.6          & 94.0          & \textbf{46.0} & 21.1          & 12.5          & \textbf{98.0} & 88.9          & 87.9          & \textbf{44.0} & \textbf{29.1} & \textbf{22.3} \\
    \quad Over + MG &  82.7          & 50.1          & 47.5          & \textbf{97.6} & \textbf{96.1} & \textbf{95.3} & \textbf{46.0} & \textbf{29.1} & \textbf{17.2} & \textbf{98.0} & \textbf{91.1} & \textbf{90.2} & 37.9          & 28.6          & 18.8 \\
    \quad Over + CG &  83.6          & \textbf{54.3} & \textbf{50.6} & - & - & - & - & - & - & - & - & - & -          & -          & - \\    
 
    \bottomrule
  \end{tabular}}
\end{table}

The results in Table \ref{tab:dev-results} show that the best configuration is \textit{overlap + coarse-grained text segments}, because it obtains the overall highest group score. Consider that for \textsc{BiSCoR-Ctrl} \textit{mid-grained} and \textit{coarse-grained} are equivalent, since we only have one attribute per object. The results indicate that processing more crops per image is beneficial. The intuition is that if we use more crops, we will better capture the objects, their attributes, and relations with other objects. On the other hand, fewer text segments but semantically richer ones offer better results. This makes sense since dual encoder VLMs are trained to process descriptive captions, and thus similarity values for shorter word segments will be generally lower. 

In addition, Table \ref{tab:dev-results} offers some interesting insights: i) both \textsc{BiSCoR} datasets are very challenging, with base VLMs below random performance; ii) our inference-time approach achieves very strong improvements (more than 42 absolute points for \textsc{BiSCoR-Nat} and up to 91 points for the best case in \textsc{BiSCoR-Ctrl}), iii) \textsc{Quantity} instances contain more than two objects, and are thus challenging for our method, which still manages to improve 13 points for SigLip 2-Giant. 

\subsection{Evaluation results}
\label{subsec:eval}
We use three existing bidirectional retrieval datasets for evaluation: i) \textsc{BiVLC} \cite{miranda2024BiVLC}, using the 359 \textsc{Swap} subset; ii) Winoground \cite{thrush2022winoground}, where we provide the results for the 171 instances that have been annotated as suitable for VL compositionality \cite{diwan2022winoground} (the results for the complete dataset can be found in Appendix~\ref{appendix:winoground_results}); iii) the test split of \textsc{BiSCoR-Ctrl}, containing 1K instances for the four attribute types constructed following the method in Section \ref{subsec:dev}. 

We evaluate the CLIP and SigLIP 2 models in their standard inference approach and with our inference-time approach working best at development (i.e. overlapping crops and coarse-grained text segments). Table \ref{tab:test-results} shows that our approach yields better group score than the baselines in 8 out of 9 cases, validating that adding simple structure at inference is effective, and that it helps improve the capabilities of dual-encoder VLMs. The results also show that the improvements come mainly from the text-to-image retrieval direction (T2I), showing that our approach manages to better balance the weaker results in this direction of the base VLMs. 

In more detail, for the two natural datasets (\textsc{BiVLC} and Winoground) the improvements are clear, specially for SigLIP 2-Giant on \textsc{BiVLC} (more than 22 points for group score), with only a slight decrease of performance for CLIP on Winoground.  
Regarding the controlled dataset, the improvements for all VLMs are very strong (23 absolute points for CLIP and up to 50 for the best SigLIP 2). Given the nature of the scene compositions in \textsc{BiSCoR-Ctrl}, we see that our inference-time approach is specially strong for attribute-object binding (see the results for all \textsc{BiSCoR-Ctrl} variants in Table \ref{tab:attr-type}). 

\begin{table}[t]
  \caption{Evaluation results. For every dataset we provide image-to-text retrieval accuracy (I2T), text-to-image retrieval accuracy (T2I) and Group score. 
  +ITA refers to our inference-time approach. 
  }
  \label{tab:test-results}
  \centering
  \resizebox{0.8\columnwidth}{!}{%
  \begin{tabular}{p{2.5cm}ccccccccccccccccccc}
    \toprule
    \multirow{2}{*}{\small\textbf{Model}} &
     \multicolumn{3}{c}{\textbf{\textsc{BiVLC}}} & \multicolumn{3}{c}{\textbf{\textsc{Winoground - 171}}} & \multicolumn{3}{c}{\textbf{\textsc{BiSCoR-Ctrl}}} \\
     \cmidrule(lr){2-4} \cmidrule(lr){5-7}  \cmidrule(lr){8-10}  
     &  \textbf{I2T} & \textbf{T2I} & \textbf{Group}& \textbf{I2T} & \textbf{T2I} & \textbf{Group}&  \textbf{I2T} & \textbf{T2I} & \textbf{Group} \\
    \midrule    
    Random & 25.0 & 25.0 & 16.7 & 25.0 & 25.0 & 16.7 & 25.0 & 25.0 & 16.7 \\
    \midrule
    CLIP  &  \textbf{46.5} & 16.2          & 13.7          & \textbf{32.2} & 11.7          & \textbf{9.4}  & 9.7           & 6.4           & 1.4 \\
    \quad + ITA  & 44.6          & \textbf{22.6} & \textbf{17.0} & 26.3          & \textbf{15.8} & 8.8           & \textbf{37.2} & \textbf{33.2} & \textbf{24.8}\\
    \midrule
     SigLIP 2  &  \textbf{55.4} & 16.4          & 13.7          & 41.0          & 12.3          & 11.7          & 26.0          & 12.0          & 6.1 \\
    \quad + ITA  &  54.6          & \textbf{29.8} & \textbf{24.5} & \textbf{43.3} & \textbf{18.7} & \textbf{17.0} & \textbf{57.7} & \textbf{59.6} & \textbf{47.9}\\
    \midrule
    SigLIP 2-Giant & \textbf{57.1} & 9.2           & 7.8           & \textbf{42.7} & 19.3         & 15.2          & 21.6          & 13.9          & 6.0 \\
    \quad + ITA  & 56.3          & \textbf{36.5} & \textbf{29.8} & 41.5          & \textbf{24.0} & \textbf{18.1} & \textbf{69.7} & \textbf{62.5} & \textbf{56.8}\\
    \bottomrule
  \end{tabular}}
\end{table}

\section{Analysis and discussion}
\label{sec:analysis}

\subsection{The importance of image crops}
\label{subsec:ablation}
As previous work has shown the benefits of decomposing the captions via specific questions about the image \cite{wiles2025gecko, hu2023tifa}, a natural question is whether image crops do actually bring any benefit to our inference-time approach. For that purpose, we perform some ablation experiments using our development datasets, \textsc{BiSCoR-Nat} and \textsc{BiSCoR-Ctrl}. The results are shown in Table \ref{tab:ablation}. As can be seen, adding \textit{coarse-grained} text segments to the base model without decomposing the image barely improves the performance and even degrades it in some cases. The largest performance improvements come when text segments are used in conjunction with image crops.

\begin{table}[t]
  \caption{Ablation results on development to measure the contribution of using text segments only, or combined with image crops for the best configurations with respec to each base model. %
  }
  \label{tab:ablation}
  \centering
  \resizebox{\columnwidth}{!}{%
  \begin{tabular}{p{2.5cm}ccccccccccccccccccc}
    \toprule
     \multirow{2}{*}{\small\textbf{Model}} &
     \multicolumn{3}{c}{\textbf{\textsc{BiSCoR-Nat}}} & \multicolumn{12}{c}{\textbf{\textsc{BiSCoR-Ctrl}}} \\
     & & & & \multicolumn{3}{c}{\textbf{\textsc{Color}}} & \multicolumn{3}{c}{\textbf{\textsc{Size}}} & \multicolumn{3}{c}{\textbf{\textsc{Material}}}& \multicolumn{3}{c}{\textbf{\textsc{Quantity}}} \\
\cmidrule(lr){2-4} \cmidrule(lr){5-7}  \cmidrule(lr){8-10}  \cmidrule(lr){11-13}  \cmidrule(lr){14-16} 
     &  \textbf{I2T} & \textbf{T2I} & \textbf{Gr.}& \textbf{I2T} & \textbf{T2I} & \textbf{Gr.}&  \textbf{I2T} & \textbf{T2I} & \textbf{Gr.} &  \textbf{I2T} & \textbf{T2I} & \textbf{Gr.}&  \textbf{I2T} & \textbf{T2I} & \textbf{Gr.} \\
    \midrule        
    Random & 25.0 & 25.0 & 16.7 & 25.0 & 25.0 & 16.7 & 25.0 & 25.0 & 16.7 & 25.0 & 25.0 & 16.7 & 25.0 & 25.0 & 16.7 \\
    \midrule
     CLIP  &  45.9          & 13.4          & 11.2          & 10.1          & 4.8           & 1.2           & 2.8           & 8.2           & 0.5         & 8.8             & 2.8          & 0.5            & \textbf{11.9}          & 7.8           & 3.5   \\
    \quad + text segments  &  38.6          & 15.5          & 12.1         & 5.6            & 5.1            & 1.1            & 14.3           & 7.2            & 2.3   
    & 0.4           & 1.8               & 0.0            & 4.8  
    & 7.6            & 1.9 \\
    \quad + image crops  &  \textbf{69.1}          & \textbf{29.5}          & \textbf{26.2}          & \textbf{84.3}          & \textbf{80.3}          & \textbf{74.9}          & \textbf{20.2}         
    & \textbf{16.3}          & \textbf{7.4}          
    & \textbf{33.0}          & \textbf{22.6}            & \textbf{12.9}          & 8.6           
    & \textbf{14.5}          & \textbf{3.6}  \\
    \midrule
     SigLIP 2  &  72.8          & 24.6          & 23.4          & 17.0          & 5.9           & 2.8           & \textbf{39.1}          & 15.9          & \textbf{8.3}           & 6.7            & 7.0          & 1.1            & 21.4          & 10.4          & 5.2 \\
    \quad + text segments  &  60.2          & 29.5          & 25.3          & 9.8            & 3.8            & 0.9            & 25.5           & 11.6          & 4.0              & 14.6 & 10.3 & 1.9 & 14.9           & 7.7   & 3.3  \\
    \quad + image crops  &  \textbf{80.3}          & \textbf{44.3}          & \textbf{41.2}          & \textbf{97.7} & \textbf{95.7}         & \textbf{95.4} & 19.5          
    & \textbf{21.6}          & 6.1           
    & \textbf{74.1}         & \textbf{82.7}             & \textbf{64.9}         & \textbf{32.6}         
    & \textbf{30.6} & \textbf{18.6}   \\
    \midrule
     SigLIP 2-Giant & 75.4          & 9.1           & 8.2           & 19.5          & 8.7           & 4.8           & 18.7          & 15.8          & 4.4           & 15.5          & 7.8            & 2.2          & 21.7          & 14.9          & 4.9   \\
    \quad + text segments  & 64.6          & 34.4          & 29.3          & 10.3           & 6.0
    & 2.8            & 22.3
    & 14.5           & 3.3            
    & 14.6 & 6.2    & 2.2 
    & 19.5           & 6.1            
    & 2.0  \\
    \quad + image crops  
    & \textbf{83.6} & \textbf{54.3} 
    & \textbf{50.6} & \textbf{97.6}   & \textbf{96.1} & \textbf{95.3} 
    & \textbf{46.0} & \textbf{29.1} 
    & \textbf{17.2} & \textbf{98.0}   & \textbf{91.1} & \textbf{90.2}   & \textbf{37.9} & \textbf{28.6}
    & \textbf{18.8}\\
 
    \bottomrule
  \end{tabular}}
\end{table}

\subsection{Analysis of the results on \textsc{BiVLC}}
\label{subsec:bivlc-analysis}

\begin{figure}[t]
    \centering 
    \begin{subfigure}[b]{0.49\linewidth}
  
         \includegraphics[width=1\textwidth]{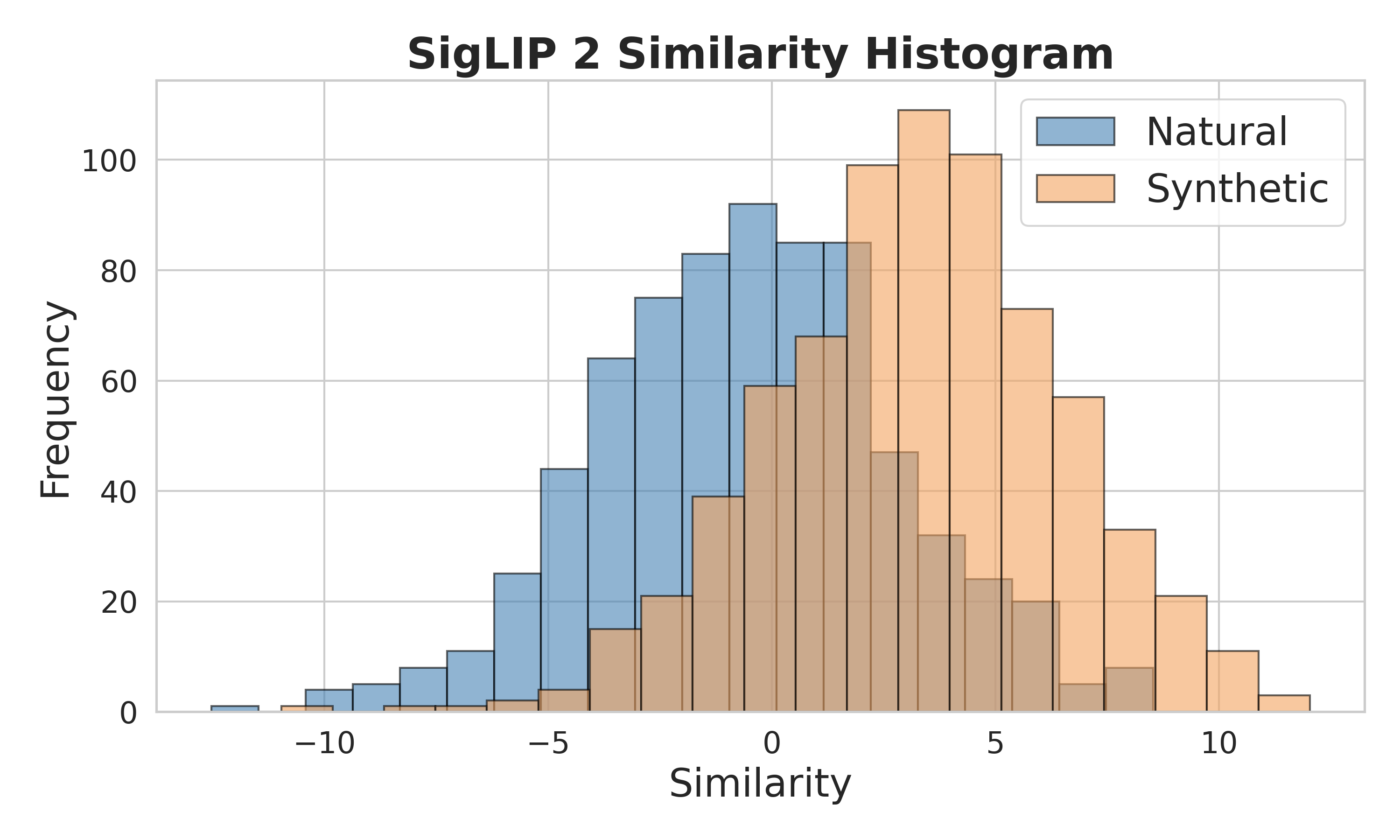}
    \end{subfigure}
    \begin{subfigure}[b]{0.49\linewidth}
         \includegraphics[width=1\textwidth]{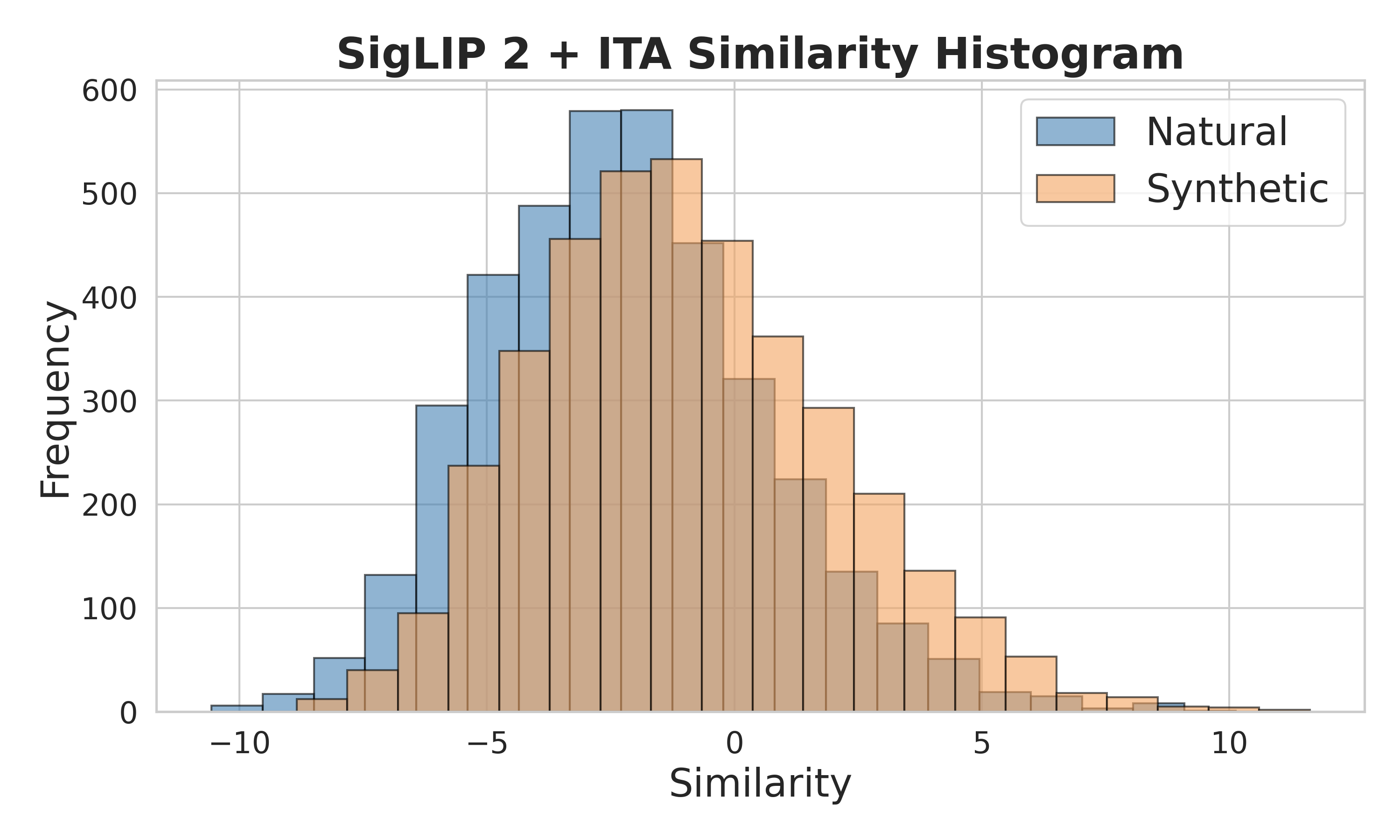}
    \end{subfigure}
    \caption{Left: the distribution of image-text similarities for SigLIP 2 base for natural and synthetic images. Right: the distribution of the similarities of the matches between image crops and text segments for natural and synthetic images for SigLIP 2 with our inference-time approach. Both histograms provide the scaled similarities (as used in SigLIP 2 implementations) calculated over \textsc{BiVLC}.}
    \label{fig:histogram-similarities}   
\end{figure}

We identify two main areas of improvement for our inference-time approach after analyzing the results on \textsc{BiVLC}:

\paragraph{Synthetic and natural images:} Each instance of \textsc{BiVLC} has a natural and a synthetic image. The synthetic image is generated from the negative caption and thus tends to clearly represent the objects and attributes mentioned in the caption. In natural images, the described objects are not often clearly visible. Furthermore, if we look at the finer metrics in \textsc{BiVLC} (Table \ref{tab:BiVLC_finer} in Appendix \ref{appendix:bivlc_finer}), we observe that base models have a much lower T\textsubscript{pos}2I accuracy than T\textsubscript{neg}2I, which means that the positive caption is frequently aligned with the negative synthetic image. Guided by those observations, we plotted two distributions (Figure \ref{fig:histogram-similarities}): the similarities of the captions for the natural and synthetic images with SigLIP 2 base model (left), and the similarities of the matches for SigLIP 2 with our approach (right). As can be seen, the similarities of the base model tend to be much higher for synthetic images, which explains the difference between T\textsubscript{pos}2I and T\textsubscript{neg}2I (over 68 points). When using our approach, the similarities of matches still tend to be higher for the synthetic images, but the bias is reduced significantly and thus the difference between T\textsubscript{pos}2I and T\textsubscript{neg}2I improves 15 points. Hence, despite our approach significantly mitigating the bias towards synthetic images of the base models, we see that our aggregation function for matches is still sensitive to this bias.

\paragraph{The generation of text segments:} As we use an LLM to generate the text segments from captions, we wanted to measure the errors introduced by this automatic process. For that purpose, we take all the instances where SigLIP 2-Giant with our inference-time approach fails (group score = 0) and manually annotate when the text segments are wrong. The heuristics to find wrong text segments are: i) different number of segments for positive and negative captions, ii) missing any object, attribute or relation and iii) hallucinating information. We find that 59\% of the text segments are wrong, accounting for an important source of error. To further validate this, we generate text segments using both the positive and negative captions together as the input of the LLM. This is not fair, since the LLM can see both captions at the same time, but we see that generally, the text segments are of higher quality. Evaluating VLMs with our inference-time approach and those \textit{higher quality} text segments, CLIP improves +1.7, SigLIP 2 +2.5 and SigLIP 2-Giant +4.2 absolute points in group score. The improvement comes mainly from I2T, suggesting that this metric is specially sensitive to the quality of text segments (see Appendix~\ref{appendix:bivlc_finer} for details). This further stresses the importance of developing better ways to generate text segments from captions.

\subsection{Analysis of the results on Winoground}
\label{subsec:wino-analysis}
We manually inspected the 171 compositional instances of Winoground and we found that many captions tend to be very brief basically naming two objects and their relation (real example: \textit{"a snake eats a bird"}). As a consequence, some of the captions are not even decomposed by our text segmenting strategies and thus our inference-time approach is not contributing anything new to the base model for those instances. For that reason, we further curated the 171 instances of Winoground to keep only those where the caption can be segmented in a meaningful way. In total, we keep 102 instances. It is interesting to see how results improve significantly for those instances: while CLIP has a modest improvement of +2.0 absolute points, both SigLIP 2 models improve +13.7 and +11.8 absolute points respectively. With those improvements, SigLIP 2 achieves a group score of 21.6 and its giant version 20.6 (see Appendix~\ref{appendix:winoground_results} for complete results). This highlights again the importance of getting good text segments and shows that our inference-time approach can actually leverage the structure of a scene as described in the textual caption. However, when the caption is too simple, e.g. no attributes, just one relation between two objects and similars, our technique is reduced to the base model's inference process.

\subsection{The influence of attribute types}
\label{subsec:attr-type}

As \textsc{BiSCoR-Ctrl} offers a variety of attribute types for compositionality, we can analyze what kinds of attributes work better with our inference-time approach. Table \ref{tab:attr-type} shows the results of our VLMs for the test split of \textsc{BiSCoR-Ctrl}. According to the results, \textsc{Color} is the easiest attribute for all the VLMs, followed by \textsc{Material}. The best model, SigLIP 2-Giant, scores over 90 for those two attributes, showing a very strong performance with our inference-time approach. However, \textsc{Size} and \textsc{Quantity} seem to be harder for all models. Although our approach brings extra performance for all the models (except for CLIP in \textsc{Quantity} and SigLIP 2 in \textsc{Size}), the results are lower than for the other attributes. As a result of a manual analysis, we saw that for \textsc{Size}, the image crops that capture objects in isolation tend to remove the size information, making hard to decide whether the object is actually large or small (see Appendix~\ref{appendix:attribute_analysis}). On the other hand, for \textsc{Quantity}, the scenes are more complex than for the other attributes with up to 10 objects. Furthermore, when we swap the quantities of two object groups to form a negative example, one of the text segments of the negative caption can find a correct match in the positive image (see Appendix~\ref{appendix:attribute_analysis}).

\begin{table}[t]
  \caption{Results for all the attribute types of the test split of \textsc{BiSCoR-Ctrl} for all the evaluated VLMs. +ITA refers to the VLM with our inference-time approach. }
  \label{tab:attr-type}
  \centering
  \resizebox{\columnwidth}{!}{%
  \begin{tabular}{p{2.5cm}cccccccccccccccccccccc}
    \toprule
    \multirow{2}{*}{\small\textbf{Model}} &
     \multicolumn{3}{c}{\textbf{\textsc{Color}}} & \multicolumn{3}{c}{\textbf{\textsc{Size}}} & \multicolumn{3}{c}{\textbf{\textsc{Material}}} & \multicolumn{3}{c}{\textbf{\textsc{Quantity}}}\\
     \cmidrule(lr){2-4} \cmidrule(lr){5-7}  \cmidrule(lr){8-10}  \cmidrule(lr){11-13} 
     &  \textbf{I2T} & \textbf{T2I} & \textbf{Group}& \textbf{I2T} & \textbf{T2I} & \textbf{Group}&  \textbf{I2T} & \textbf{T2I} & \textbf{Group} & \textbf{I2T} & \textbf{T2I} & \textbf{Group} \\
    \midrule    
    Random & 25.0 & 25.0 & 16.7 & 25.0 & 25.0 & 16.7 & 25.0 & 25.0 & 16.7 & 25.0 & 25.0 & 16.7 \\
    \midrule
     CLIP  &  9.6           & 6.1           & 1.5           & 2.9           & 7.0           & 0.1           & 11.6          & 2.8           & 0.7           & \textbf{14.6} & 9.8           & \textbf{3.4} \\
    \quad + ITA  & \textbf{85.6} & \textbf{80.6} & \textbf{76.0} & \textbf{17.7} & \textbf{13.0} & \textbf{5.1}  & \textbf{38.3} & \textbf{26.4} & \textbf{16.0} & 7.0           & \textbf{12.9} & 1.9   \\
    \midrule
     SigLIP 2  & 20.3          & 7.8           & 3.6           & \textbf{42.3} & 16.5          & \textbf{7.9}  & 8.4           & 7.1           & 1.5          & 33.1          & 16.5          & 11.3 \\
    \quad + ITA  & \textbf{97.9} & \textbf{94.8} & \textbf{94.4} & 20.20          & \textbf{25.3} & 5.9           & \textbf{78.4} & \textbf{82.2} & \textbf{68.6} & \textbf{34.3} & \textbf{36.2} & \textbf{22.8}\\
    \midrule
    SigLIP 2-Giant & 20.8          & 12.0          & 6.0           & 23.0          & 18.4          & 6.8           & 15.8          & 7.8           & 2.3           & 26.6          & 17.5          & 9.0 \\
    \quad + ITA  & \textbf{97.8} & \textbf{96.4} & \textbf{95.7} & \textbf{49.1} & \textbf{30.5} & \textbf{20.2} & \textbf{98.2} & \textbf{92.4} & \textbf{91.7} & \textbf{33.6} & \textbf{30.6} & \textbf{19.4}\\
    \bottomrule
  \end{tabular}}
\end{table}

\section{Conclusions}

We have presented a new approach that tries to tackle compositionality issues of current dual encoder VLMs at inference time. Our experiments show that structuring the image as overlapping crops of varying shapes and sizes, coupled with structuring the text as text segments improves performance on both controlled experiments on automatically created synthetic data as well as natural images and captions. %
We focus our evaluation on \textsc{Swap} cases, where both captions of a retrieval instance are composed of the same words but attributes are interchanged, as they show the limitations of current dual encoders to model compositions of objects and attributes.  %

As future work, we would like to explore more sophisticated structures for both images and text, including object and relation detections in images as well as more accurate text segmentation methods. In addition we would like to explore also better aggregation functions for the final similarity score.
\newpage

{
\small

\bibliography{bibliography.bib}
}

\newpage
\appendix

\section{Limitations and societal impact}

\subsection{Limitations}\label{appendix:limitations}
During the analysis of the results, we did not explore variations for the following:  i) the generation of text segments from natural captions, and ii) the method to aggregate scores. Exploring alternative techniques for those two components could improve further the results. Another limitation is the computational cost of the current implementation, as it needs multiple inferences per image-text pair. However, this can be optimized too, mimicking object detectors such as Faster R-CNN and similars \cite{girshick2015fast}, which use ROI Align to represent various regions of an image in a single inference. Further research could explore how to integrate those techniques in our inference pipeline. Finally, we only considered English to evaluate VL compositionality. This is mainly because there are not VL compositionality datasets in other languages. As SigLIP 2 models are already multilingual, it would be interesting to see how they perform for other languages.

\subsection{Societal impacts}\label{appendix:SOCIETAL}

Vision-language models such as CLIP are popular models for many applications, but previous research has probed and analyzed their limitations\citep{yuksekgonul2022and, hsieh2024sugarcrepe, miranda2024BiVLC}. We contribute with our research by delving into a new viewpoint for improving the compositionality of dual encoder models: adding structure at inference-time. We hope that it will encourage the improvement of compositionality in inference, opening the way to more complex algorithms that better exploit the compositional capabilities of the models and impacting on image-text retrieval systems, which are currently of common use among many people.

\section{Detailed evaluation metrics}
\label{appendix:metrics}
The \textbf{I2T} score measures the performance for image-to-text retrieval. For each instance in our dataset, we actually have two image-to-text retrieval examples. To obtain a perfect I2T score, the correct captions for both images have to be selected. Thus, assuming $C_0, C_1$ refer to positive and negative caption respectively, $I_0, I_1$ to positive and negative image, and we use $s(C_i, I_i)$ as the similarity function for a caption and an image, I2T score $I2T(C_0, I_0, C_1, I_1)$ is defined in Equation \ref{eq:text-score}: 
     \begin{equation}
     \label{eq:text-score}
    I2T\left(C_{0}, I_{0}, C_{1}, I_{1}\right)=\left\{\begin{array}{cc}
     1 & \textnormal { if } s\left(C_{0}, I_{0}\right)>s\left(C_{1}, I_{0}\right)\\
    & \textnormal { and } s\left(C_{1}, I_{1}\right)>s\left(C_{0}, I_{1}\right) \\
     0 & \textnormal { otherwise }
    \end{array}\right.
    \end{equation}
    
The \textbf{T2I} score $T2I(C_{0}, I_{0}, C_{1}, I_{1})$ is similarly defined for text-to-image retrieval (Equation \ref{eq:img-score}):

    \begin{equation}
    \label{eq:img-score}
    T2I\left(C_{0}, I_{0}, C_{1}, I_{1}\right)=\left\{\begin{array}{cc}
     1 & \textnormal { if } s\left(C_{0}, I_{0}\right)>s\left(C_{0}, I_{1}\right)  \\
    & \textnormal { and } s\left(C_{1}, I_{1}\right)>s\left(C_{1}, I_{0}\right) \\
     0 & \textnormal { otherwise }
    \end{array}\right.
    \end{equation}

Finally, the \textbf{Group} score $G(C_{0}, I_{0}, C_{1}, I_{1})$ is the main metric, since it combines the performance for image-to-text and text-to-image retrieval. To obtain a perfect group score for a given instance, both images have to be matched with the suitable captions and both captions with the suitable images. The group score is defined in Equation \ref{eq:group-score}:

    \begin{equation}    
    \label{eq:group-score}
    G\left(C_{0}, I_{0}, C_{1}, I_{1}\right)=\left\{\begin{array}{cc}
    1 & \textnormal { if } I2T\left(C_{0}, I_{0}, C_{1}, I_{1}\right)\\
    & \textnormal { and } T2I\left(C_{0}, I_{0}, C_{1}, I_{1}\right) \\
    0 & \textnormal { otherwise }
    \end{array}\right.
    \end{equation}

\section{\textsc{BiSCoR} dataset information}\label{appendix:dataset_info}

We host \textsc{BiSCoR} at HuggingFace\footnote{https://huggingface.co/datasets/imirandam/BiSCoR}. We provide a summary below.

\paragraph{Dataset documentation} \textsc{BiSCoR} is a benchmark of Bidirectional \textsc{Swaps} for Compositional Reasoning development.
Each instance consists of two images and two captions. Using each of the images and captions as a base, a model is asked to select the pair that correctly represents the base versus the hard negative distractor with minor compositional changes. Thus, we can measure image-to-text and text-to-image retrieval with hard negative pairs. To obtain good results on the dataset, it is necessary that the model performs well in both directions for the same instance. It's formed by two main branches, \textsc{BiSCoR-Nat} and \textsc{BiSCoR-Ctrl}.

\textsc{BiSCoR-Nat} is a curated version of \textsc{TROHN-Img} \citep{miranda2024BiVLC} with 427 high-quality \textsc{Swaps}. Each instance of the dataset consists of four fields

\begin{itemize}
    \item image: COCO 2017 train image.
    \item caption: COCO 2017 train text describing the COCO image.
    \item negative\_caption: Negative caption generated from the COCO text description by \citep{miranda2024BiVLC}.
    \item negative\_image: Negative image generated from the negative caption by \citep{miranda2024BiVLC}.
\end{itemize}

\textsc{BiSCoR-Ctrl} is designed to have full control of the scenes and their compositions, allowing for a more detailed assessment of compositionality. It is based on CLEVR \citep{johnson2017clevr}, where we build different variants of \textsc{Swap} instances: i) \textsc{Color}, ii) \textsc{Size}, iii) \textsc{Material} and iv) \textsc{Quantity}. Each variant consists of two splits, Development and Test, with 1000 instances each.

\begin{itemize}
    \item image: New positive image rendered by us.
    \item caption: Caption obtained from the scene used to render the positive image.
    \item negative\_image: New negative image rendered by us
    \item negative\_image: Caption obtained from the scene used to render the negative images.
\end{itemize}

\begin{figure}[h]
  \centering \includegraphics[scale=0.53]{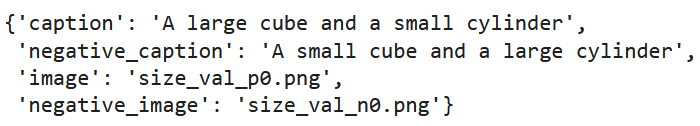}
  \caption{Example of a \textsc{BiSCoR} instance after loading the dataset.}
  \label{fig:instance_example}
\end{figure}

\paragraph{Maintenance plan} We are committed to maintaining the dataset to resolve any technical issues. We actively track issues in the HuggingFace or GitHub repositories. %
\paragraph{Licensing} Our work is licensed under the MIT License\footnote{https://github.com/IMirandaM/structure-inference-compositionality/blob/main/LICENSE} for the code and a Creative Commons Attribution 4.0 International License (CC BY 4.0) for the data\footnote{https://huggingface.co/datasets/choosealicense/licenses/blob/main/markdown/cc-by-4.0.md}. 
\paragraph{Author statement} We, the authors, assume full responsibility in case of violation of rights.

\section{Implementation details}\label{appendix:implementation}

This appendix contains all the information related to the implementation of the experiments. All the source code with instructions can be found at \url{https://github.com/IMirandaM/structure-inference-compositionality}. %

\subsection{Source datasets}\label{appendix:source}
We obtain all source datasets directly from the original sources published by the authors. To the best of our knowledge, all data sources we use are open to non-commercial use, do not contain personally identifiable information and do not contain offensive content.

\begin{itemize}
    \item \textbf{CLEVR} \citep{lin2014microsoft}: We obtain CLEVR scenes from the official project website\footnote{\url{https://cs.stanford.edu/people/jcjohns/clevr/}} under a Creative Commons Attribution 4.0 License. 
    \item \textbf{TROHN-Img} \citep{miranda2024BiVLC}: We obtain the swap instances from the official Hugging Face repository \footnote{\url{https://huggingface.co/datasets/imirandam/TROHN-Img}} under the MIT license. 
\end{itemize}

\subsection{Evaluation datasets}\label{appendix:eval_data}

We obtain all evaluation datasets directly from the original sources published by the authors.

\begin{itemize}
    \item \textbf{\textsc{BiVLC}} \citep{miranda2024BiVLC}: We obtain \textsc{BiVLC} \textsc{Swap} instances from the official Hugging Face repository \footnote{\url{https://huggingface.co/datasets/imirandam/BiVLC}}. 
    \item \textbf{\textsc{Winoground}} \citep{yuksekgonul2022and}: We obtain \textsc{Winoground} from the official Hugging Face repository \footnote{\url{https://huggingface.co/datasets/facebook/winoground}}.
\end{itemize}

\subsection{Software information}\label{appendix:software}
\paragraph{Models}
We detail the sources of models we used.

\begin{itemize}
    \item \textbf{CLIP} We obtain the pretrained baseline VIT-B-32 OpenAI's CLIP model \citep{radford2021learning} from Hugging Face \footnote{\url{https://huggingface.co/openai/clip-vit-base-patch32}}.
   
    \item \textbf{SigLIP 2} We obtain all SigLIP 2\citep{tschannen2025siglip} models from Hugging Face collection\footnote{\url{https://huggingface.co/collections/google/siglip2-67b5dcef38c175486e240107}}.
    \begin{itemize}
        \item \textbf{SigLIP 2}: We obtain
siglip2-base-patch32-256 from the official Hugging Face repository\footnote{\url{https://huggingface.co/google/siglip2-base-patch32-256}}.
        \item \textbf{SigLIP 2-Giant}: We obtain
siglip2-giant-opt-patch16-256 from the official Hugging Face repository\footnote{\url{https://huggingface.co/google/siglip2-giant-opt-patch16-256}}.
    \end{itemize}

\end{itemize}

\paragraph{Implementation decisions}

We have decided to keep the preprocessing of the images constant based on the model with the lowest resolution, i.e. CLIP, controlling that all models receive the same original image and same crops. For this, all images are preprocessed in the same way, resize to 224 and center crop. All hyperparameters described in Section \ref{sec:experiments} are identical across all models. The only difference is the batch size: 25 for CLIP and SigLIP 2, and 5 for SigLIP 2-Giant.

\paragraph{Evaluation} We base our evaluations on Transformers library.

\paragraph{Rendering images} For rendering the images from \textsc{BiSCoR-Ctrl} we have used Blender 4.4.3.

\subsection{Hardware information}\label{appendix:hardware}

\paragraph{Development experiments} All development experiments have been performed on one NVIDIA A100-SXM4-80GB GPU and 128 GB of RAM. In terms of computation time, to execute the 6 configurations over 1000 instances takes less than one hour for CLIP, 1.5 hours for SigLIP 2 and 14 hours for SigLIP 2-Giant. 

\paragraph{Evaluation} As in the development experiments, the evaluation has been performed on one NVIDIA A100-SXM4-80GB GPU and 128 GB of RAM. In terms of computation time, for \textsc{BiVLC} and \textsc{Winoground}: CLIP 5 seconds, CLIP + ITA 8 minutes, SigLIP 2 6 seconds, SigLIP 2 + ITA 12 minutes, SigLIP 2-Giant 35 seconds and SigLIP 2-Giant + ITA 2 hours. For \textsc{BiSCoR-Ctrl}: CLIP 36 seconds, CLIP + ITA 1.33 hours, SigLIP 2 45 seconds, SigLIP 2 + ITA 2 hours, SigLIP 2-Giant 6 minutes and SigLIP 2-Giant + ITA 21 hours.

\paragraph{Rendering images} For rendering the images we have used an NVIDIA RTX A1000 6GB Laptop GPU. Each rendering takes around 1.5 seconds. 

\subsection{LLM for text segments generation} \label{appendix:LLM}

To generate text segments from natural captions we used the Gemini 2.0 Flash model using the official Gemini API \footnote{\url{https://ai.google.dev/gemini-api/docs/models}}. We used the next hyperparameters:
\begin{itemize}
    \item temperature : 0.0
    \item top\_k : 1
    \item top\_p : None
\end{itemize}

All the templates can be found in Appendix~\ref{appendix:templates}

\section{Real examples of \textsc{Size} and \textsc{Quantity} errors }\label{appendix:attribute_analysis}
\subsection{Real example of \textsc{Size} attribute error}
\begin{figure}[h]
    \centering 
         \includegraphics[width=0.75\textwidth]{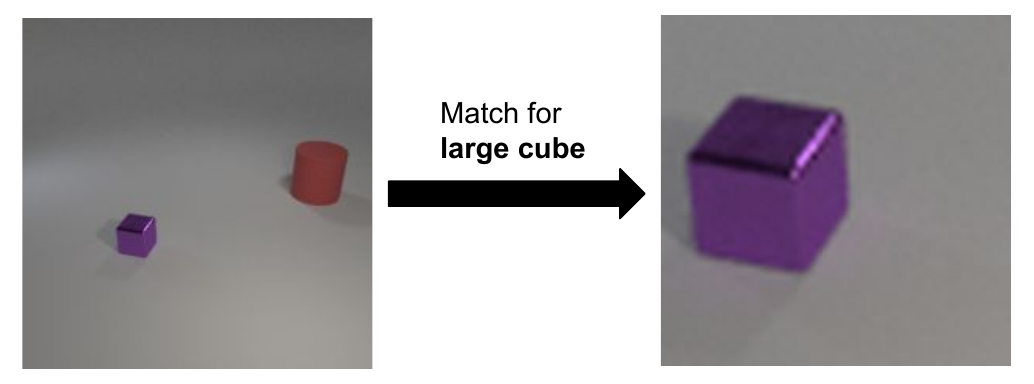}   
    \caption{A real example where given the positive image and negative text segments "small cylinder", "large cube" and "small cylinder and a large cube", the match of "large cube" is an image crop that captures the object in isolation removing the size information.}
    \label{fig:size_analysis}   
\end{figure}

\subsection{Real example of \textsc{Quantity} attribute error}
\begin{figure}[h]
    \centering 
         \includegraphics[width=0.75\textwidth]{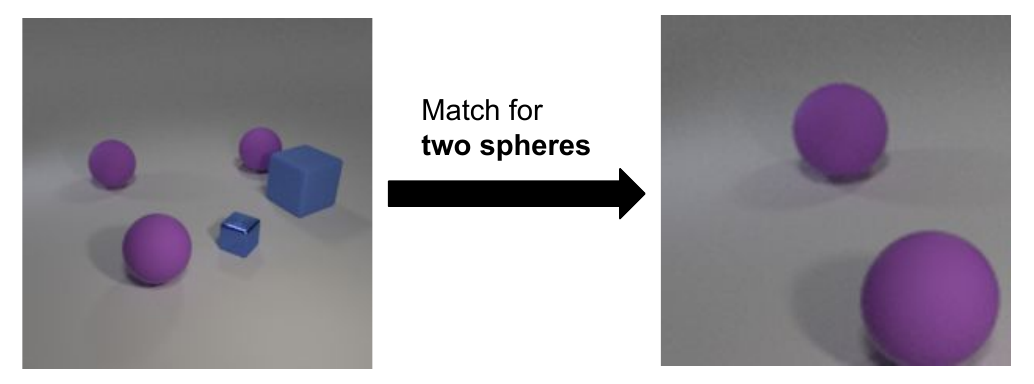}   
    \caption{A real example where given the positive image and negative text segments "three spheres", "two cubes" and "three spheres and two cubes", the match is an image crop that captures correctly the information from the text segments, "two spheres", but not the global information of the image, "three spheres".}
    \label{fig:quantity_analysis}   
\end{figure}

\newpage
\section{Development Results}
\label{appendix:dev_results}

In this section, we present the development results of the CLIP and SigLIP 2 models.

\subsection{CLIP development results}

\begin{table}[ht]
  \caption{Development results for CLIP base model and different inferece-time approach configurations. FG for \textit{fine-grained}, MG for \textit{mid-grained} and CG for \textit{coarse-grained} text segments. Note that for \textsc{BiSCoR-Ctrl} the MG and CG text segments are the same. Bold for best results. }
  \label{tab:dev-clip}
  \centering
  \resizebox{\columnwidth}{!}{%
 \begin{tabular}{p{2.5cm}ccccccccccccccccccc}
    \toprule
     \multirow{2}{*}{\small\textbf{Model}} &
     \multicolumn{3}{c}{\textbf{\textsc{BiSCoR-Nat}}} & \multicolumn{12}{c}{\textbf{\textsc{BiSCoR-Ctrl}}} \\
     & & & & \multicolumn{3}{c}{\textbf{\textsc{Color}}} & \multicolumn{3}{c}{\textbf{\textsc{Size}}} & \multicolumn{3}{c}{\textbf{\textsc{Material}}}& \multicolumn{3}{c}{\textbf{\textsc{Quantity}}} \\
\cmidrule(lr){2-4} \cmidrule(lr){5-7}  \cmidrule(lr){8-10}  \cmidrule(lr){11-13}  \cmidrule(lr){14-16} 
     &  \textbf{I2T} & \textbf{T2I} & \textbf{Gr.}& \textbf{I2T} & \textbf{T2I} & \textbf{Gr.}&  \textbf{I2T} & \textbf{T2I} & \textbf{Gr.} &  \textbf{I2T} & \textbf{T2I} & \textbf{Gr.}&  \textbf{I2T} & \textbf{T2I} & \textbf{Gr.} \\
    \midrule    
    Random & 25.0 & 25.0 & 16.7 & 25.0 & 25.0 & 16.7 & 25.0 & 25.0 & 16.7 & 25.0 & 25.0 & 16.7 & 25.0 & 25.0 & 16.7\\
    \midrule
    CLIP  & 45.9          & 13.4          & 11.2          & 10.1          & 4.8           & 1.2           & 2.8           & 8.2           & 0.5          & 8.8           & 2.8           & 0.5           & \textbf{11.9} & 7.8           & 3.5  \\
    \midrule
    \quad Grid + FG & 60.9          & 22.7          & 19.2          & 69.4          & 60.5          & 52.9          & 18.0          & 8.9           & 3.6          & 23.7          & 13.7          & 6.9           & 8.5           & 11.5          & 2.5 \\
    \quad Grid + MG &  61.1          & 29.7          & 25.5          & 69.4          & 67.1          & 57.4          & 18.0          & 15.5          & 6.5          & 23.7          & 16.6          & 7.8           & 6.9           & 10.7          & 2.1  \\
    \quad Grid + CG &  62.1          & 29.5          & 24.6          & -          & -         & -         & -          & -          & -          & -         & -          & -           & -          & -          & -  \\
    \midrule
    \quad Over + FG & 65.1          & 22.5          & 19.0          & \textbf{84.3} & 74.1          & 69.5          & \textbf{20.2} & 10.4          & 4.8          & \textbf{33.0} & 19.3          & 10.3          & 10.2          & \textbf{16.4} & \textbf{4.7} \\
    \quad Over + MG &  67.2          & \textbf{30.4} & \textbf{26.2} & \textbf{84.3} & \textbf{80.3} & \textbf{74.9} & \textbf{20.2} & \textbf{16.3} & \textbf{7.4} & \textbf{33.0} & \textbf{22.6} & \textbf{12.9} & 8.6           & 14.5          & 3.6 \\
    \quad Over + CG &  \textbf{69.1} & 29.5          & \textbf{26.2} & - & - & - & - & - & - & - & - & - & - & - & -\\    
 
    \bottomrule
  \end{tabular}}
\end{table}

\subsection{SigLIP 2 development results}

\begin{table}[ht]
  \caption{Development results for SigLIP 2 base model and different inferece-time approach configurations. FG for \textit{fine-grained}, MG for \textit{mid-grained} and CG for \textit{coarse-grained} text segments. Note that for \textsc{BiSCoR-Ctrl} the MG and CG text segments are the same. Bold for best results. }
  \label{tab:dev-siglip}
  \centering
  \resizebox{\columnwidth}{!}{%
 \begin{tabular}{p{2.5cm}ccccccccccccccccccc}
    \toprule
     \multirow{2}{*}{\small\textbf{Model}} &
     \multicolumn{3}{c}{\textbf{\textsc{BiSCoR-Nat}}} & \multicolumn{12}{c}{\textbf{\textsc{BiSCoR-Ctrl}}} \\
     & & & & \multicolumn{3}{c}{\textbf{\textsc{Color}}} & \multicolumn{3}{c}{\textbf{\textsc{Size}}} & \multicolumn{3}{c}{\textbf{\textsc{Material}}}& \multicolumn{3}{c}{\textbf{\textsc{Quantity}}} \\
\cmidrule(lr){2-4} \cmidrule(lr){5-7}  \cmidrule(lr){8-10}  \cmidrule(lr){11-13}  \cmidrule(lr){14-16} 
     &  \textbf{I2T} & \textbf{T2I} & \textbf{Gr.}& \textbf{I2T} & \textbf{T2I} & \textbf{Gr.}&  \textbf{I2T} & \textbf{T2I} & \textbf{Gr.} &  \textbf{I2T} & \textbf{T2I} & \textbf{Gr.}&  \textbf{I2T} & \textbf{T2I} & \textbf{Gr.} \\
    \midrule      
    Random & 25.0 & 25.0 & 16.7 & 25.0 & 25.0 & 16.7 & 25.0 & 25.0 & 16.7 & 25.0 & 25.0 & 16.7 & 25.0 & 25.0 & 16.7\\
    \midrule
    SigLIP 2  & 72.8          & 24.6          & 23.4          & 17.0          & 5.9           & 2.8           & \textbf{39.1} & 15.9          & \textbf{8.3} & 6.7           & 7.0           & 1.1           & 21.4          & 10.4          & 5.2 \\
    \midrule
    \quad Grid + FG & 79.9          & 33.5          & 31.4          & 96.3          & 89.4          & 88.4          & 17.8          & 12.4          & 4.1          & 68.8          & 63.4          & 49.6          & 35.0          & 28.1          & 17.7 \\
    \quad Grid + MG &  79.6          & 41.0          & 38.9          & 96.3          & 92.3          & 91.3          & 17.8          & 14.7          & 5.1          & 68.8          & 71.6          & 55.4          & 30.7          & 28.0          & 17.1  \\
    \quad Grid + CG &  80.1          & 41.5          & 39.1          & -          & -          & -          & -          & -          & -          & -          & -          & -          & -          & -          & - \\
    \midrule
    \quad Over + FG & \textbf{82.0} & 35.4          & 33.3          & \textbf{97.7} & 94.8          & 94.5          & 19.5          & 18.5          & 5.7          & \textbf{74.1} & 78.4          & 62.3          & \textbf{37.3} & \textbf{31.9} & \textbf{21.3} \\
    \quad Over + MG &  81.7          & \textbf{44.3} & 41.0          & \textbf{97.7} & \textbf{95.7} & \textbf{95.4} & 19.5          & \textbf{21.6} & 6.1          & \textbf{74.1} & \textbf{82.7} & \textbf{64.9} & 32.6          & 30.6          & 18.6  \\
    \quad Over + CG &  80.3          & \textbf{44.3} & \textbf{41.2} & - & - & - & -          & - & -          & - & - & - & -         & -          & - \\    
 
    \bottomrule
  \end{tabular}}
\end{table}

\newpage
\section{Winoground Complete Results}
\label{appendix:winoground_results}

\begin{table}[h]
  \caption{Results of the VLMs for Winoground, in terms of image-to-text retrieval (I2T), text-to-image retrieval (T2I) and Group score. +ITA refers to the VLM with our inference-time approach. We provide the results of a random system as reference. As can be seen, the group score of the VLMs with our inference-time approach improves consistently.}
  \label{tab:wino-whole-results}
  \centering
  \resizebox{1\columnwidth}{!}{%
  \begin{tabular}{p{2.5cm}ccccccccc}
    \toprule
    \multirow{2}{*}{\small\textbf{Model}} &
     \multicolumn{3}{c}{\textbf{\textsc{Winoground}}} & \multicolumn{3}{c}{\textbf{\textsc{Winoground - 171}}} & \multicolumn{3}{c}{\textbf{\textsc{Winoground - 102}}}\\
     \cmidrule(lr){2-4} \cmidrule(lr){5-7}  \cmidrule(lr){8-10}
     &  \textbf{I2T} & \textbf{T2I} & \textbf{Group}& \textbf{I2T} & \textbf{T2I} & \textbf{Group} & \textbf{I2T} & \textbf{T2I} & \textbf{Group} \\
    \midrule    
    Random & 25.0 & 25.0 & 16.7 & 25.0 & 25.0 & 16.7 & 25.0 & 25.0 & 16.7 \\
    \midrule
    CLIP  &  \textbf{31.25} & 11.00          & \textbf{8.75}  & \textbf{32.16} & 11.70          & \textbf{9.36} & \textbf{28.4} & 7.8           & 6.9 \\
    \quad + ITA  & 23.75          & \textbf{12.00} & 6.75           & 26.32          & \textbf{15.79} & 8.77 & 27.5          & \textbf{17.7} & \textbf{8.8} \\
    \midrule
     SigLIP 2  &  \textbf{37.25} & 13.25          & 10.75          & 40.94          & 12.28          & 11.70 & 38.2          & 7.8           & 7.8 \\
    \quad + ITA  &  36.00          & \textbf{18.00} & \textbf{13.75} & \textbf{43.27} & \textbf{18.71} & \textbf{16.96} & \textbf{49.0} & \textbf{24.5} & \textbf{21.6}\\
    \midrule
    SigLIP 2-Giant & \textbf{38.75} & 17.00          & 13.25          & \textbf{42.69} & 19.30          & 15.20 & 37.3          & 12.8          & 8.8  \\
    \quad + ITA  & 36.25          & \textbf{21.25} & \textbf{14.75} & 41.52          & \textbf{23.98} & \textbf{18.13} & \textbf{47.1} & \textbf{24.5} & \textbf{20.6}\\
    \bottomrule
  \end{tabular}}
\end{table}

\section{BiVLC results with finer-metrics and higher quality text segments}\label{appendix:bivlc_finer}

\begin{table}[h]
  \caption{Results of the VLMs on \textsc{BiVLC}, in terms of image-to-text retrieval (I2T), text-to-image retrieval (T2I) and Group score, plus individual retrieval task scores: I\textsubscript{pos}2T refers to image-to-text retrieval with positive image, I\textsubscript{neg}2T for image-to-text retrieval with negative image, etc.. +ITA refers to the VLM with our inference-time approach and  +ITA-HQ refers to the VLM with our inference-time approach with high quality text segments. We provide the results of a random system as reference. As can be seen, T\textsubscript{pos}2I is much lower than T\textsubscript{neg}2I for all the cases, showing a bias to select the negative image. On the other hand, +ITA-HQ mainly improves I2T, suggesting that this metric is particularly sensitive to the quality of text segment}
  \label{tab:BiVLC_finer}
  \centering
  \resizebox{1\columnwidth}{!}{%
  \begin{tabular}{p{2.5cm}cccccccc}
    \toprule
    \multirow{2}{*}{\textbf{Model}} & \multicolumn{3}{c}{\textbf{\textsc{BiVLC}}} & \multicolumn{4}{c}{\textbf{\textsc{BiVLC} (finer metrics)}}\\ 
    \cmidrule(lr){2-4} \cmidrule(lr){5-8}
     & \textbf{I2T} & \textbf{T2I} & \textbf{Group} & \textbf{I\textsubscript{pos}2T} & \textbf{I\textsubscript{neg}2T} & \textbf{T\textsubscript{pos}2I} & \textbf{T\textsubscript{neg}2I} \\
    \midrule    
    Random & 25.0 & 25.0 & 16.7 & 50.0 & 50.0 & 50.0 & 50.0 \\
    \midrule
    CLIP  & 46.5 & 16.2          & 13.7          & 66.0          & \textbf{72.4} & \textbf{43.5} & 71.9  \\
    \quad + ITA  & 44.6          & 22.6 & 17.0 & 72.7 & 64.9          & 42.3          & \textbf{78.0}\\
    \quad + ITA-HQ & \textbf{53.8} & \textbf{22.8} & \textbf{18.7} & \textbf{75.2} & 71.9          & 42.9          & 77.4  \\
    \midrule
     SigLIP 2  &  55.4 & 16.4          & 13.7          & 73.3          & \textbf{77.7} & 23.1          & \textbf{91.9} \\
    \quad + ITA  &  54.6          & 29.8 & 24.5 & 78.8 & 72.1          & 37.1 & 90.5 \\
    \quad + ITA-HQ & \textbf{61.3} & \textbf{30.9} & \textbf{27.0} & \textbf{80.8} & 76.3 & \textbf{38.7} & 90.3          \\
    \midrule
    SigLIP 2-Giant  & 57.1 & 9.2           & 7.8           & 76.3          & 74.4          & 10.9          & \textbf{97.8} \\
    \quad + ITA  & 56.3          & 36.5 & 29.8 & 76.6 & 75.2 & \textbf{49.6} & 85.2 \\
    \quad + ITA-HQ & \textbf{65.2} & \textbf{37.6} & \textbf{34.0} & \textbf{81.3}          & \textbf{81.1}          & \textbf{49.6}          & 87.5 \\
    \bottomrule
  \end{tabular}}
\end{table}

\newpage
\section{\textsc{BiSCoR} instance examples}\label{appendix:more_examples}

\begin{figure}[h]
  \centering \includegraphics[width=1\textwidth]{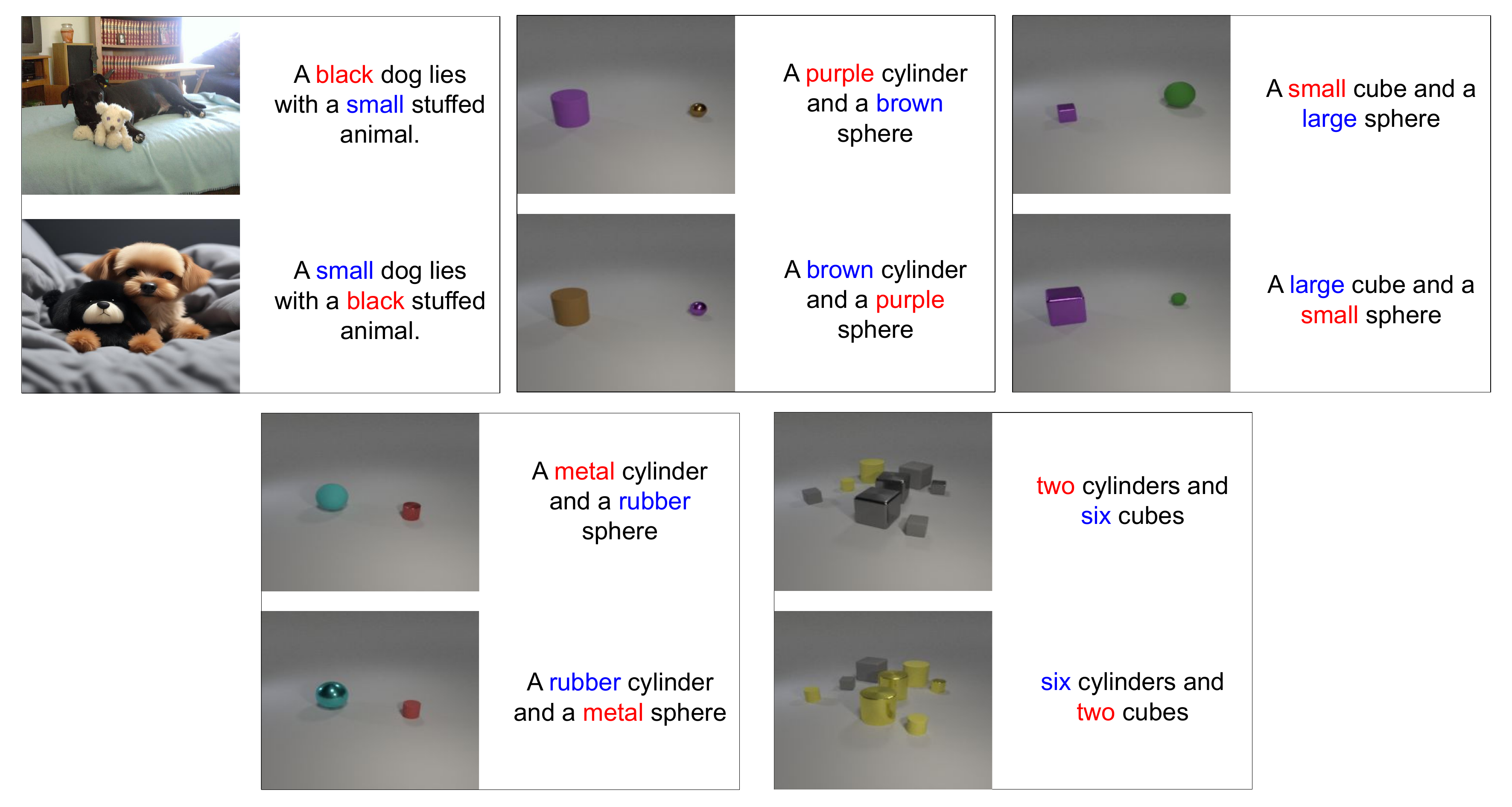}
  \caption{Five instances from the \textsc{BiSCoR} development dataset. First row from left-to-right an instance from \textsc{BiSCoR-Nat} followed by two instances of \textsc{BiSCoR-Ctrl}, the first one for \textsc{Color} and the second for \textsc{Size}, in the second row from left-to-right an instance of \textsc{Material} and an instance of \textsc{Quantity} with 8 objects. Each instance consists of a positive pair (top image and caption) and a hard negative pair (bottom image and caption).}
  \label{fig:more-examples}
\end{figure}

\newpage
\section{Templates}\label{appendix:templates}

In this section, we show the different templates used during experimentation with different text segments granularities.

\subsection{Coarse-grained template}

\begin{listing}[!h]
\begin{minted}[breaklines,obeytabs=true,tabsize=2,breaksymbolleft=]{python}
'''Construct concise compositional text segments...

For captions with multiple attributes for a single object, the text segment must contain the object and all its attributes. (e.g., "small white toilet").

For captions with multiple relations for a single object, include text segments for each object-relation combination and aggregations (e.g., "white toilet with black seat", "white toilet with silver handle", "white toilet with black seat and silver handle").

It is important that for relations text segments the objects are ALWAYS accompanied by all their attributes.

It is IMPORTANT to use the same words and structures present in the caption.

It is IMPORTANT that the final text segments fully represent the provided caption, from the details to the more global context.

This is the caption: "A white toilet with a black seat."

"text_segments": [
    "white toilet",
    "black seat",
    "white toilet with black seat",
]

This is the caption: "A fluffy white cat sleeping on a brown rug."

"text_segments": [
    "fluffy white cat",
    "brown rug",
    "fluffy white cat sleeping",
    "fluffy white cat sleeping on brown rug"
]

This is the caption: '''
\end{minted}
\caption{Coarse-grained template}
\label{listing:coarse_template}
\end{listing}

\newpage
\subsection{Mid-grained template}
\begin{listing}[!h]
\begin{minted}[breaklines,obeytabs=true,tabsize=2,breaksymbolleft=]{python}
'''Construct concise compositional text segments...

For captions with multiple attributes for a single object, include text segments for each object-attribute combination and aggregations (e.g., "white toilet", "small toilet", "small white toilet").

For captions with multiple relations for a single object, include text segments for each object-relation combination and aggregations (e.g., "white toilet with black seat", "white toilet with silver handle", "white toilet with black seat and silver handle").

It is important that for relations text segments the objects are ALWAYS accompanied by all their attributes.

It is IMPORTANT to use the same words and structures present in the caption.

It is IMPORTANT that the final text segments fully represent the provided caption, from the details to the more global context.

This is the caption: "A white toilet with a black seat."

"text_segments": [
    "white toilet",
    "black seat",
    "white toilet with black seat",
]

This is the caption: "A fluffy white cat sleeping on a brown rug."

"text_segments": [
    "white cat",
    "fluffy cat",
    "fluffy white cat",
    "brown rug",
    "fluffy white cat sleeping",
    "fluffy white cat sleeping on brown rug"
]

This is the caption: '''
\end{minted}
\caption{Mid-grained template}
\label{listing:mid_template}
\end{listing}

\newpage
\subsection{Fine-grained template}

\begin{listing}[!h]
\begin{minted}[breaklines,obeytabs=true,tabsize=2,breaksymbolleft=]{python}
'''Construct concise compositional text segments...

For captions with multiple objects, create separate sets of text segments focusing on each main object individually (e.g. "toilet", "seat").

For captions with multiple attributes for a single object, include text segments for each object-attribute combination and aggregations (e.g., "white toilet", "small toilet", "small white toilet").

For captions with multiple relations for a single object, include text segments for each object-relation combination and aggregations (e.g., "white toilet with black seat", "white toilet with silver handle", "white toilet with black seat and silver handle").

It is important that for relations text segments the objects are ALWAYS accompanied by all their attributes.

It is IMPORTANT to use the same words and structures present in the caption.

It is IMPORTANT that the final text segments fully represent the provided caption, from the details to the more global context.

This is the caption: "A white toilet with a black seat."

"text_segments": [
    "toilet",
    "white toilet",
    "seat",
    "black seat",
    "white toilet with black seat",
]

This is the caption: "A fluffy white cat sleeping on a brown rug."

"text_segments": [
    "cat",
    "white cat",
    "fluffy cat",
    "fluffy white cat",
    "rug",
    "brown rug",
    "fluffy white cat sleeping",
    "fluffy white cat sleeping on brown rug"
]

This is the caption: '''
\end{minted}
\caption{Fine-grained template}
\label{listing:fine_template}
\end{listing}

\newpage
\section*{NeurIPS Paper Checklist}

\begin{enumerate}

\item {\bf Claims}
    \item[] Question: Do the main claims made in the abstract and introduction accurately reflect the paper's contributions and scope?
    \item[] Answer: \answerYes{} %
    \item[] Justification: The main claims in the abstract and introduction provide a clear summary of the main conclusions and are aligned with the experiments performed and the results obtained in the paper.

\item {\bf Limitations}
    \item[] Question: Does the paper discuss the limitations of the work performed by the authors?
    \item[] Answer: \answerYes{} %
    \item[] Justification: See Appendix~\ref{appendix:limitations}

\item {\bf Theory assumptions and proofs}
    \item[] Question: For each theoretical result, does the paper provide the full set of assumptions and a complete (and correct) proof?
    \item[] Answer: \answerNA{} %
    \item[] Justification: We do not present any theoretical results.

    \item {\bf Experimental result reproducibility}
    \item[] Question: Does the paper fully disclose all the information needed to reproduce the main experimental results of the paper to the extent that it affects the main claims and/or conclusions of the paper (regardless of whether the code and data are provided or not)?
    \item[] Answer:\answerYes{} %
    \item[] Justification: The method proposed in the paper is explained in detail and models, data, hyperparameters and computational resources are provided. All code, data, etc. necessary to reproduce the experiments will be published.

\item {\bf Open access to data and code}
    \item[] Question: Does the paper provide open access to the data and code, with sufficient instructions to faithfully reproduce the main experimental results, as described in supplemental material?
    \item[] Answer: \answerYes{} %
    \item[] Justification: All code and data will be public in the official GitHub repository of the project. This repository contains detailed instructions and resources (code, data, etc.) to reproduce all the experiments performed in the paper.

\item {\bf Experimental setting/details}
    \item[] Question: Does the paper specify all the training and test details (e.g., data splits, hyperparameters, how they were chosen, type of optimizer, etc.) necessary to understand the results?
    \item[] Answer: \answerYes{} %
    \item[] Justification: See Section~\ref{sec:experiments}

\item {\bf Experiment statistical significance}
    \item[] Question: Does the paper report error bars suitably and correctly defined or other appropriate information about the statistical significance of the experiments?
    \item[] Answer: \answerNo{} %
    \item[] Justification: The dual encoder models and algorithms used in our experiments are deterministic.

\item {\bf Experiments compute resources}
    \item[] Question: For each experiment, does the paper provide sufficient information on the computer resources (type of compute workers, memory, time of execution) needed to reproduce the experiments?
    \item[] Answer: \answerYes{} %
    \item[] Justification: See Appendix~\ref{appendix:hardware} 
    
\item {\bf Code of ethics}
    \item[] Question: Does the research conducted in the paper conform, in every respect, with the NeurIPS Code of Ethics \url{https://neurips.cc/public/EthicsGuidelines}?
    \item[] Answer: \answerYes{} %
    \item[] Justification: We have read it and agree.

\item {\bf Broader impacts}
    \item[] Question: Does the paper discuss both potential positive societal impacts and negative societal impacts of the work performed?
    \item[] Answer: \answerYes{} %
    \item[] Justification: See Appendix~\ref{appendix:SOCIETAL}
    
\item {\bf Safeguards}
    \item[] Question: Does the paper describe safeguards that have been put in place for responsible release of data or models that have a high risk for misuse (e.g., pretrained language models, image generators, or scraped datasets)?
    \item[] Answer: \answerNA{} %
    \item[] Justification: The data we publish does not contain any risk of misuse. It comes from secure sources and has been filtered by us. 

\item {\bf Licenses for existing assets}
    \item[] Question: Are the creators or original owners of assets (e.g., code, data, models), used in the paper, properly credited and are the license and terms of use explicitly mentioned and properly respected?
    \item[] Answer: \answerYes{} %
    \item[] Justification: All the information (license, URLs, etc.) about source datasets, evaluation datasets and models can be found in Appendix~\ref{appendix:implementation}. 

\item {\bf New assets}
    \item[] Question: Are new assets introduced in the paper well documented and is the documentation provided alongside the assets?
    \item[] Answer: \answerYes{} %
    \item[] Justification: All the information regarding the development dataset can be found in Appendix~\ref{appendix:dataset_info}. Some data, such as DOI, metadata, URLs, etc., are omitted so as not to break anonymity. It will be included in the camera-ready version. 

\item {\bf Crowdsourcing and research with human subjects}
    \item[] Question: For crowdsourcing experiments and research with human subjects, does the paper include the full text of instructions given to participants and screenshots, if applicable, as well as details about compensation (if any)? 
    \item[] Answer: \answerNA{} %
    \item[] Justification: We have not performed any crowdsourcing process.

\item {\bf Institutional review board (IRB) approvals or equivalent for research with human subjects}
    \item[] Question: Does the paper describe potential risks incurred by study participants, whether such risks were disclosed to the subjects, and whether Institutional Review Board (IRB) approvals (or an equivalent approval/review based on the requirements of your country or institution) were obtained?
    \item[] Answer:\answerNA{} %
    \item[] Justification: \answerNA{}

\item {\bf Declaration of LLM usage}
    \item[] Question: Does the paper describe the usage of LLMs if it is an important, original, or non-standard component of the core methods in this research? Note that if the LLM is used only for writing, editing, or formatting purposes and does not impact the core methodology, scientific rigorousness, or originality of the research, declaration is not required.
    \item[] Answer: \answerNA{} %
    \item[] Justification: \answerNA{}

\end{enumerate}

\end{document}